\documentclass[lettersize,journal]{IEEEtran}
\usepackage{amsmath,amsfonts}
\usepackage{algorithmic}
\usepackage{algorithm}
\usepackage{array}
\usepackage{textcomp}
\usepackage{stfloats}
\usepackage{url}
\usepackage{verbatim}
\usepackage{graphicx}
\usepackage{cite}
\usepackage{booktabs}
\usepackage{multirow}
\usepackage[caption=false,font=scriptsize,labelfont=rm,textfont=rm]{subfig}

\usepackage{subfloat}
\usepackage[pagebackref=true,breaklinks=true,letterpaper=true,colorlinks,bookmarks=false]{hyperref}
\usepackage[capitalize]{cleveref}
\begin{document}

\title{Relationship Quantification of Image Degradations}

\author{%
    Wenxin Wang,
    Boyun Li,
    Yuanbiao Gou,
    Peng Hu,
    Wangmeng Zuo, 
    and Xi Peng
    \\
\thanks{W. Wang, B. Li, Y. Gou, P. Hu and X. Peng are with College of Computer Science, Sichuan University, Chengdu, 610065, China.
 e-mail: \{wangwenxin.gm, liboyun.gm, gouyuanbiao, penghu.ml, pengx.gm\}@gmail.com.}
\thanks{M. Zuo is with School of Computer Science and Technology, Harbin Institute of Technology, Harbin, 150001, China.
 e-mail: wmzuo@hit.edu.cn}
\thanks{W. Wang and B. Li have contribute equally to this work.}
\thanks{Corresponding author: X. Peng. }
}

\markboth{Journal of \LaTeX\ Class Files,~Vol.~14, No.~8, August~2021}%
{Shell \MakeLowercase{\textit{et al.}}: A Sample Article Using IEEEtran.cls for IEEE Journals}


\maketitle

\begin{abstract}
In this paper, we study two challenging but less-touched problems in image restoration, namely, i) how to quantify the relationship between image degradations and ii) how to improve the performance of a specific restoration task using the quantified relationship. To tackle the first challenge, we proposed a Degradation Relationship Index (DRI) which is defined as the mean drop rate difference in the validation loss between two models which are respectively trained using the anchor degradation and the mixture of the anchor and the auxiliary degradations. Through quantifying the degradation relationship using DRI, we reveal that i) a positive DRI always predicts performance improvement by using the specific degradation as an auxiliary to train models; ii) the degradation proportion is crucial to the image restoration performance. In other words, the restoration performance is improved only if the anchor and the auxiliary degradations are mixed with an appropriate proportion. Based on the observations, we further propose a simple but effective method (dubbed DPD) to estimate whether the given degradation combinations could improve the performance on the anchor degradation with the assistance of the auxiliary degradation. Extensive experimental results verify the effectiveness of our method in dehazing, denoising, deraining, and desnowing. The code will be released after acceptance.
\end{abstract}

\begin{IEEEkeywords}
Image dehazing; Auxiliary degradation based restoration; Image restoration.
\end{IEEEkeywords}

\section{Introduction}

In the real world, images are often contaminated by various degradations such as noise, rain, haze, and snow, thus deteriorating the imaging quality and making difficulty in image restoration. To obtain human-favorite images, plentiful works have been conducted and significant developments have been achieved during past years~\cite{DnCNN, DeblurGAN, JORDER, RESCAN, MPRNet, SwinIR, Restormer, pan2013, TMICS, MTCRN, AGSN,RNAN,RDN}.

Although these methods perform well by specifically designing a model for a specific degradation (referring to as one-for-one), they are less attractive to some real-world scenarios such as autopilot which highly expect to tackle multiple degradations using a unified model. To develop such an all-in-one model, some studies have been conducted for blind~\cite{AirNet,MCR} or non-blind restoration~\cite{AIO, TransWeather, IPT} by designing a new network architecture, objective function, or training strategy. Although promising results have been achieved, these works focus on removing multiple degradations from inputs, which however largely ignore another important and promising topic, \textit{i.e.},  Relationship Quantification of Image Degradations (RQID). Notably, RQID does not serve as a metric to directly measure the similarity between two degradations due to its abstract nature and immeasurability. Instead, it serves as a quantification tool to reflect the positive or negative influence of the auxiliary degradation on the anchor restoration task, so that the performance of the anchor restoration could be improved as desired. 

\begin{figure}[t]
    \centering
    \includegraphics[scale=0.65]{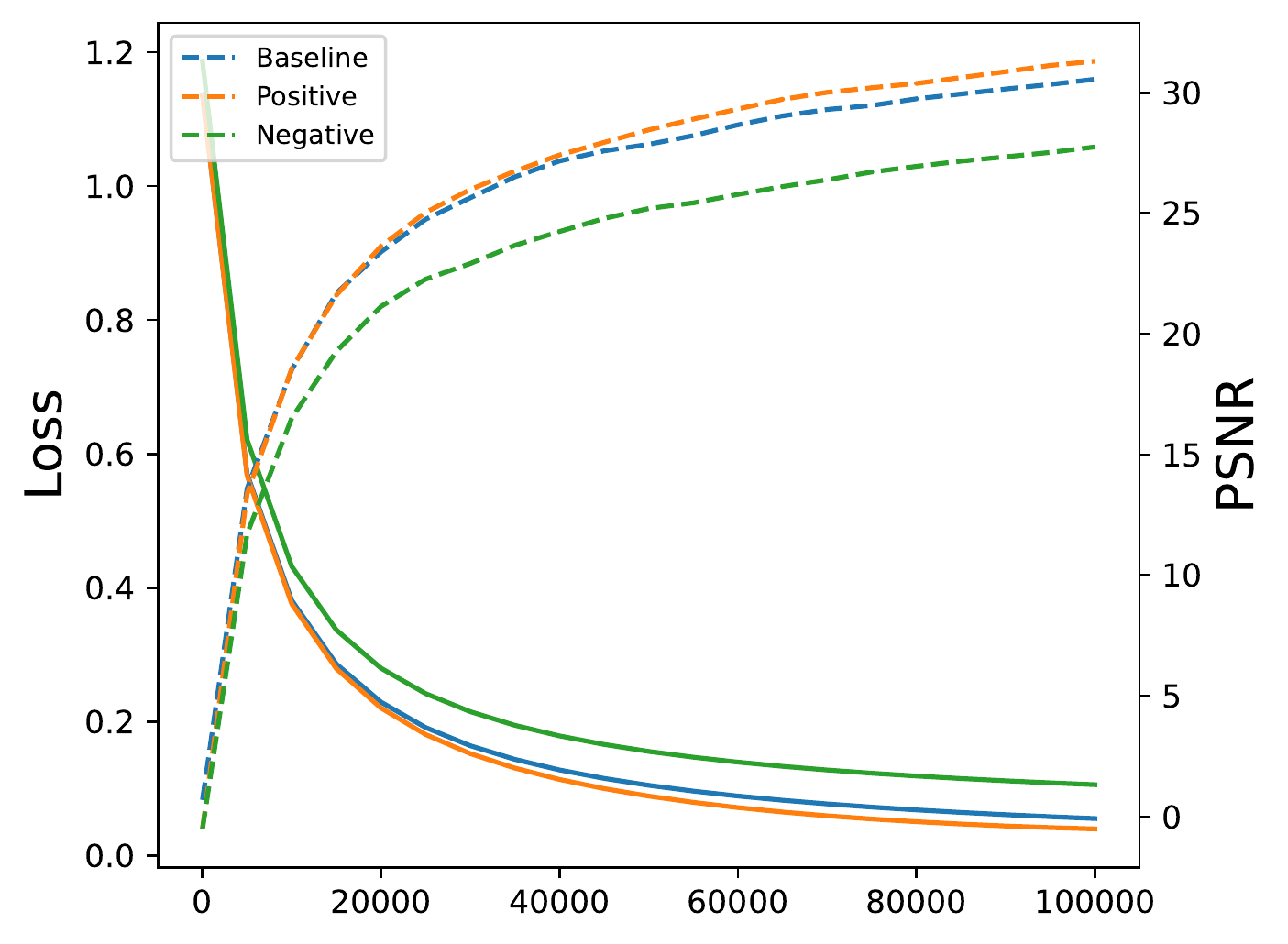}
   \caption{Our observation and the effectiveness of DRI. As a baseline, the blue solid curve illustrates the validation loss of the model trained on the anchor degradation (\textit{e.g.}, haze). Achieving better and worse restoration performance respectively, the red and green solid curves are the validation losses of the models trained with the anchor and the auxiliary degradations (\textit{e.g.}, noise), where the only difference between them is the proportion of the auxiliary degradation.}
   \label{Fig:observation}
\end{figure}

To implement RQID, in this paper, we propose Degradation Relationship Index (DRI) and Degradation Proportion Determination method (DPD) to quantitatively answer whether a given restoration task such as dehazing will be beneficial from adding another degradation (e.g., noisy images) into training procedure. In brief, DRI, which is proposed to quantify the relationship of the anchor and the auxiliary degradations, is defined by the mean drop rate difference in the validation loss between two models, where one model is trained with the anchor degradation and the other is trained with both the anchor and the auxiliary degradations. As shown in~\cref{Fig:observation}, a positive DRI always predicts the performance improvement of image restoration, which will prove to be correct in the following analysis. Back to the figure, one could see that compared with the baseline (blue curve), the model with the positive DRI (red curve) converges faster and achieves better performance, indicating the positive influence of the auxiliary degradation. In contrast,  the model with the negative DRI (green curve) converges slower and performs worse, indicating the negative influence of the auxiliary degradation. Interestingly, the only difference between the models with red and green curves lies in the proportion of the auxiliary degradation. Hence, one could conclude that the degradation combination, including the degradation type and proportion, is crucial to the image restoration performance. In other words, the performance of the anchor restoration will be improved only if the degradations are combined with appropriate proportions. Based on this observation and using DRI as a metric, we show that DPD, a simple but effective method would seek a desirable degradation combination to improve the performance of the anchor restoration.

Notably, DRI is different from so-called task affinity~\cite{Taskonomy, TAG} in given aspects. On the one hand, DRI is compatible with all losses and networks as it is agnostic to model and training criteria, thus enjoying flexibility. In contrast, almost all existing task affinity methods highly rely on their elaborately-designed network structure. On the other hand,  DRI could indicate whether the degradation combination boosts the original model, thus embracing predictability. In contrast, the task affinity methods do not enjoy such characteristics in image restoration as verified in~\cref{Sec:TA}.

To summarize, the contribution and novelty of this study are as below:
\begin{itemize}[itemsep=2pt,topsep=1pt,parsep=1pt]
    \item To the best of our knowledge, this work could be the first attempt on exploring and exploiting the relationship between different image degradations. 
    \item The proposed DRI can quantify the relationship between two given degradations, which enjoys the merits of predictability and flexibility. In a quantitative manner, DRI could measure the influence of a given degradation on another, \textit{i.e.}, whether the anchor restoration will be beneficial from using a specific degradation as auxiliary. 
    \item Due to the unavailability in the benchmark of this newly-emerged topic, we build a new benchmark, dubbed RESIDE+. Thanks to RESIDE+, the RQID methods could be largely immune to the influence of content discrepancy and thus specialize in the relationship of different degradations. 
    Extensive experiments show the flexibility and effectiveness of our method which remarkably improves the performance of seven representative baselines. 
\end{itemize}


\section{Related Work}
This section will briefly review recent developments in two related topics, namely, the all-in-one restoration (multiple degradations restoration, MDR) and image deweathering.

\subsection{Multiple Degradations Restoration}
To handle multiple degradations commonly seen in real-world scenarios, a large number of MDR methods~\cite{AirNet, MCR, AIO, TransWeather, IPT, Liu:2019uh} have been proposed to recover clean images from the degraded observations which contain different degradations. Although significant developments have been achieved by these studies, most of them focus on removing multiple degradations using a single model by designing new network architectures, objective functions or training strategies, which ignore a latent RQID topic in the area. 

Different from the aforementioned studies on MDR, this work does not attempt to develop a new image restoration method, which instead aims to develop a RQID tool. As a result, this study could i) provide an innovative mathematical tool to measure the interaction of different degradations, and ii) improve the performance of the anchor restoration task by introducing an auxiliary restoration task. 

\subsection{Image Deweathering}
Image deweathering aims to remove adverse weather, \textit{e.g.}, rain streak, haze and snow, to achieve appealing results for users and downstream models, \textit{e.g.}, classification~\cite{ResNet,Vit}, detection~\cite{YOLO,RetinaNet} and segmentation~\cite{Mask_R_CNN,Deeplab}, which trained on clear weather conditions. Without loss of generality, we take image dehazing, deraining and desnowing as examples to investigate the RQID.

Image dehazing aims to restore the scene radiance of the observed hazy images. With the development of deep learning, image dehazing has achieved promising results~\cite{AOD-Net, ZID, YOLY, GridDehazeNet, FFA-Net, GCA-Net, MSBDN, AECR-Net, dehamer,LHPP,UHD}. In recent, the focus of the community has shifted to designing a high-efficiency network or task-specific loss function. For instance, Dong~\textit{et~al.}~\cite{MSBDN} exploited the multi-scale information of hazy images by designing a pyramid-like network. 
Wu~\textit{et~al.}~\cite{AECR-Net} proposed a novel contrastive regularization as the loss term to leverage connections among recovered images, hazy images and ground truth. 

Similar to image dehazing, image deraining~\cite{BRN,PReNet} and desnowing~\cite{desnownet,HDCW-Net} aim to remove the effect of the rain streak and snow from the degraded images. Most of them are focusing on task-specific network structuring and loss function designing, etc, and achieve promising results.  

Different from the existing methods, our work does not attempt to improve the restoration performance by resorting to designing a new network structure, loss function, or optimization strategy. Instead, this study shows another feasible but ignored way, \textit{i.e.}, introducing an auxiliary restoration task to improve the generalization of models. As aforementioned, such a task is nontrivial due to the abstract nature and immeasurability of degradations. Besides, the difficulty also comes from the seemingly random results as shown in \Cref{Fig:observation}. The proposed RQID solution enjoys the following two highly-expected merits. On the one hand, it is network- and criterion-agnostic and thus enjoys high generalizability and flexibility. In other words, it could improve the performance of most existing image deweathering models, \textit{e.g.}, image dehazing models. On the other hand, it is free from extra training costs, thus it embraces computational efficiency.
\section{Proposed Method}
First, we will elaborate on several important design principles for quantifying degradation relationship in~\cref{Sec:Principles}. Then, based on the principles, DRI is proposed to measure the degradation relationship in~\cref{Sec:DRI}. After that,~\cref{Sec:DPD} presents a  Degradation Proportion Determination strategy (DPD) to estimate and improve the performance on the anchor degradation. At last, we will show the details on how to build the RESIDE+ benchmark in~\cref{Sec:RESIDE+}.

\subsection{Principles of Quantifying Degradation Relationship}
\label{Sec:Principles}
RQID is an innovative topic in low-level computer vision, and everything about it keeps unknown. To quantify the degradations relationship correctly and robustly, we propose the following two principles as guidance for developing our method, namely, the asymmetric and content irrelevant principles.

\textbf{Asymmetric principle.} 
Different from the similarity metric which generally embraces the symmetric property, \textit{e.g.}, Euclidean distance and structure similarity, the quantified relationship between degradations should be asymmetric since RQID aims to measure the influence of the auxiliary degradation on the anchor restoration. Clearly, such a goal is asymmetric. In other words, a  degradation combination is positive to the anchor restoration, which might be negative to the auxiliary restoration. 


\textbf{Content irrelevant principle.} 
Another designing principle to RQID is content irrelevant, \textit{i.e.}, the RQID metric will only reflect the relationship between degradations and be immune to the influence of image content. In practice, however, it is daunting even impossible to fully follow this principle since the degradations are coupled with the image content and it is nearly impossible to separate them. Hence, as a  remedy, one could carry out experiments on degraded images that are with the same image content and different degradation types. Considering degradation is the only difference among them, the quantification results could be regarded as content irrelevant approximately.

\subsection{Degradation Relationship Index}
\label{Sec:DRI}
In this section, we elaborate on the proposed DRI which quantifies the relationship between the auxiliary and the anchor degradations through the mean drop rate difference $\bar{D}$ in validation loss between two models, where one is trained to handle the anchor degradation and the other is trained to handle the two degradations. 

To compute DRI, we need to calculate the drop rate of validation loss on two models, where one is trained to handle the anchor degradation and the other is trained to handle the anchor and auxiliary degradations. To be specific, for a given batch of training samples $\mathcal{X}^{1, 2}=\{ X^1, X^2\}=\{x_1^1,\cdots\, x_i^1, x_{i+1}^2, \cdots\, x_N^2\}$ with batch size $N$, let $X^1$, $X^2$ and $\mathcal{X}^{1, 2}$ denote the samples with the anchor degradation, auxiliary degradation, and both of them, respectively. 
Moreover, for the given validation data batches $\mathcal{X}_v^{1}$ and $\mathcal{X}_v^{1,2}$, we use $\mathcal{L}(X_v^{1}, \theta^t)$ and $\mathcal{L}(\mathcal{X}_v^{1,2}, \theta^t)$ to denote the corresponding validation loss with the parameters $\theta^t$ at step $t$. 

Given the training batch $\mathcal{X}^{1,2}$, without loss of generality, the stochastic gradient descent is adopted to update the model $\theta^t$, then
\begin{align}
    \theta^{t+1}_{\mathcal{X}^{1,2}} &= \theta^t - \eta \nabla \mathcal{L}(\mathcal{X}^{1,2}, \theta^t),
    \label{theta1}
\end{align}
where $\eta$ is the learning rate and $\theta^{t+1}_{\mathcal{X}^{1, 2}}$ is the parameter set at step $t+1$  with respect to the input $\mathcal{X}^{1,2}$. 
With the model $\theta^{t+1}_{\mathcal{X}^{1,2}}$, the validation loss w.r.t. $X^{1}_{v}$ could be computed by $\mathcal{L}(X_v^1, \theta^{t+1}_{\mathcal{X}^{1,2}})$.

Similarly, suppose $\mathcal{X}^{1}=\{x_1^1,\cdots x_N^1\}$ denote the training batch that only contains the anchor degradation, 
then the model $\theta^{t+1}_{\mathcal{X}^1}$ is updated with $\mathcal{X}^{1}$ by
\begin{align}
    \theta^{t+1}_{\mathcal{X}^1} &= \theta^t - \eta \nabla \mathcal{L}(\mathcal{X}^1, \theta^t).
    \label{theta2}
\end{align}
Accordingly, the validation loss w.r.t. $X_{v}^1$ could be written as $\mathcal{L}(X_v^1,\theta^{t+1}_{\mathcal{X}^1})$.

Letting $D_t$ denote the drop rate difference at time step $t$, it gives that
\begin{equation}
\begin{aligned}
D_t & = \Phi(\mathcal{X}^{1},X_v^1,t) - \Phi(\mathcal{X}^{1,2},X_v^1,t) \\
& =\frac{\mathcal{L}(X_v^1, \theta_{\mathcal{X}^1}^{t+1}) - \mathcal{L}(X_v^1, \theta_{\mathcal{X}^{1,2}}^{t+1})}{\mathcal{L}(X_v^1, \theta^t)},
\end{aligned}
\label{Eq:DRI}
\end{equation}
where $\Phi(\mathcal{X}^{1},X_v^1,t)$ and  $\Phi(\mathcal{X}^{1,2},X_v^1,t)$ are the drop rates of the validation loss on the two aforementioned models.

With the drop rate difference, DRI is defined by the averaged $D_t$ across the training stage to mitigate volatility of drop rate difference. Mathematically, 
\begin{equation}
    \Bar{D} = \frac{1}{T} \sum_{t=1}^{T} D_t,
    \label{Eq:Relationship}
\end{equation}
where $T$ is the maximal training step.

From the definition of DRI, a positive $\Bar{D}$ indicates that the model trained with the auxiliary degradation will lead to a higher drop rate than that trained with the anchor degradation. In other words, the given degradation combination could improve the performance on the anchor restoration task if $\Bar{D}$ is with positive values. 

\begin{table}[ht]
    \centering
    \small
    \caption{Performances on image dehazing with different proportions of auxiliary restoration task, \textit{i.e.}, image denoising. Results in boldface indicate the setting with the best performance.
    }
    \begin{tabular}{c | c c c c c}
        \toprule
        $P$ &    DRI      & PSNR  & $\Delta_{PSNR}$ & SSIM & $\Delta_{SSIM}$  \\
        \midrule
        0\%   &  0        & 33.84 &    -  & 0.9849 & -      \\
    \bf{10\%} &  \bf{0.00090}  & \bf{33.96} &  \bf{0.12} & \bf{0.9850}  &  \bf{0.0001}\\
        30\%  & -0.00301  & 33.00 & -0.84 & 0.9828  & -0.0021\\
        50\%  & -0.01222  & 32.81 & -1.03 & 0.9827  & -0.0022\\
        70\%  & -0.02784  & 32.10 & -1.74 & 0.9804  & -0.0045\\
        90\%  & -0.05140  & 30.96 & -2.88 & 0.9764  & -0.0085\\
        \bottomrule
    \end{tabular}
    \label{Tab:ratio1}
\end{table}

\subsection{Degradation Proportion Determination}
\label{Sec:DPD}
Thanks to DRI, we could quantitatively analyze the relationship between different degradations. As a showcase, we take MSBDN~\cite{MSBDN}, one of the representative image dehazing methods to investigate how the performance on the anchor restoration task (\textit{i.e.}, image dehazing) is influenced by the auxiliary restoration task (\textit{i.e.}, image denoising). The proportions of the auxiliary degradation $P$ vary from 0.1 to 0.9 with a gap of 0.2.

From \cref{Tab:ratio1}, ones could obtain the following observations: i) the degradation combination proportion is crucial to the image restoration performance. In other words, the combinations with only appropriate degradation proportions, \textit{e.g.}, 10\% noise, are beneficial to the anchor restoration. While, other cases damage the performance on the anchor restoration task; ii) DRI is highly related to the restoration performance on the anchor task. In brief, a positive DRI always predicts the performance improvement of image restoration, while a negative DRI predicts a decrease in performance.

Based on the above observations, we propose the Degradation Proportion Determination strategy (DPD) to estimate and improve the performance of the anchor restoration task by seeking a desirable auxiliary proportion so that DRI is positive. The implement details of DPD is summarized in~\cref{alg1}.

DPD enjoys the following merits. On the one hand, DPD is a time-saving strategy to estimate and determine given proportions whether improve the performance on the anchor restoration task or not. By reducing the sampling interval of DRI, the DPD could enjoy $3.33\times$ speedup and thus obtain high economy. On the other hand, DPD embraces high interpretability, which makes its results explainable and truthful.

\begin{algorithm}[t!]
    \caption{Degradation Proportion Determination} 
    \label{alg1} 
    \begin{algorithmic}
        \REQUIRE A given dataset $\mathcal{D}$, auxiliary degradation proportion $r$, network parameter $\theta$, batch size $N$, maximal training step $T$.
        \FOR {$t = 1$ to $T$}
        \STATE Sample two mini-batches data $\mathcal{X}^{1}$, $\mathcal{X}^{2}$ from $\mathcal{D}$
        \STATE Mix $\mathcal{X}^{1}$ and $\mathcal{X}^{2}$ with proportion $r$ and obtain $\mathcal{X}^{1,2}$
        \STATE Update the network parameter through \cref{theta1,theta2} and get $\theta_{\mathcal{X}^{1,2}}^{t+1}, \theta_{\mathcal{X}^1}^{t+1}$.
        \STATE Calculate $D_{t}$ through \cref{Eq:DRI}.
        \ENDFOR
        \STATE Calculate DRI via $\Bar{D} = \frac{1}{T}\sum_{t=1}^{T} D_t$.
        \IF{$ \Bar{D} > 0$}
        \STATE Given degradation combination with proportion $r$, it would improve the performance on anchor restoration task.
        \ELSE
        \STATE Given degradation combination with proportion $r$, it would damage the performance on anchor restoration task.
        \ENDIF
    \end{algorithmic} 
\end{algorithm}

\subsection{Benchmark}
\label{Sec:RESIDE+}
As mentioned in~\cref{Sec:Principles}, it is highly expected to eliminate the content discrepancy so that an accurate degradation relationship could be measured. To this end, each clean image should have multiple degraded versions in experiments. As there is none of the existing datasets could satisfy the requirements. Hence, we expand the well-known image dehazing dataset, \textit{i.e.}, RESIDE~\cite{RESIDE}, and obtain its multiple degradation versions for evaluations. The new benchmark is termed RESIDE+ which contains 10, 931 clean images with their corresponding rainy, hazy, and snowy versions.

\begin{table}[ht]
    \centering
    \small
    \caption{Number of samples in each subset of RESIDE+.}
    \begin{tabular}{c|c|c c c}
        \toprule
              & clean & haze & rain & snow \\
        \midrule
         ITS+ & 1399  & 13990 & 13990 & 13990 \\
         OTS+ & 8970  & 313950 & 313950 & 313950 \\
         SVS (indoor)  & 10   & 10 & 10 & 10 \\
         SVS (outdoor)  & 10  & 10 & 10 & 10 \\
         SOTS+ (indoor) & 50 & 500 & 500 & 500\\
         SOTS+ (outdoor) & 492 & 500 & 500 & 500 \\
        \bottomrule
    \end{tabular}
    \label{Tab:RESIDE+}
\end{table}

Similar to~\cite{RESIDE}, RESIDE+ consists of four subsets, \textit{i.e.}, the Indoor Training Set+ (ITS+), Outdoor Training Set+ (OTS+), Synthetic Validation Set (SVS) and Synthetic Objective Testing Set+ (SOTS+). As shown in~\cref{Tab:RESIDE+}, ITS+ consists of 13,990 synthetic rainy, hazy and snowy images corresponding to 1, 399 clean indoor images gathered from NYUv2~\cite{NYU2} and MiddleBury stereo dataset~\cite{middle}. OTS+ contains 313, 950 rainy, hazy, and snowy outdoor images with 8, 970 clear indoor images collected from the Internet. 
SVS is a novel synthesized validation set, which consists of indoor and outdoor subsets, each of which includes 10 clean images and 10 rainy, hazy and snowy images, respectively. Clean images of indoor scenes are collected from the NYUv2~\cite{NYU2} and outdoor clean images are gathered from the Hybrid Subjective Testing Set (HSTS) of RESIDE. SOTS+ consists of indoor and outdoor test sets, and each of them contains 50 clean images, and 500 rainy, hazy and snowy images respectively.

\begin{figure}[t!]
    \centering
    \includegraphics[scale=0.13]{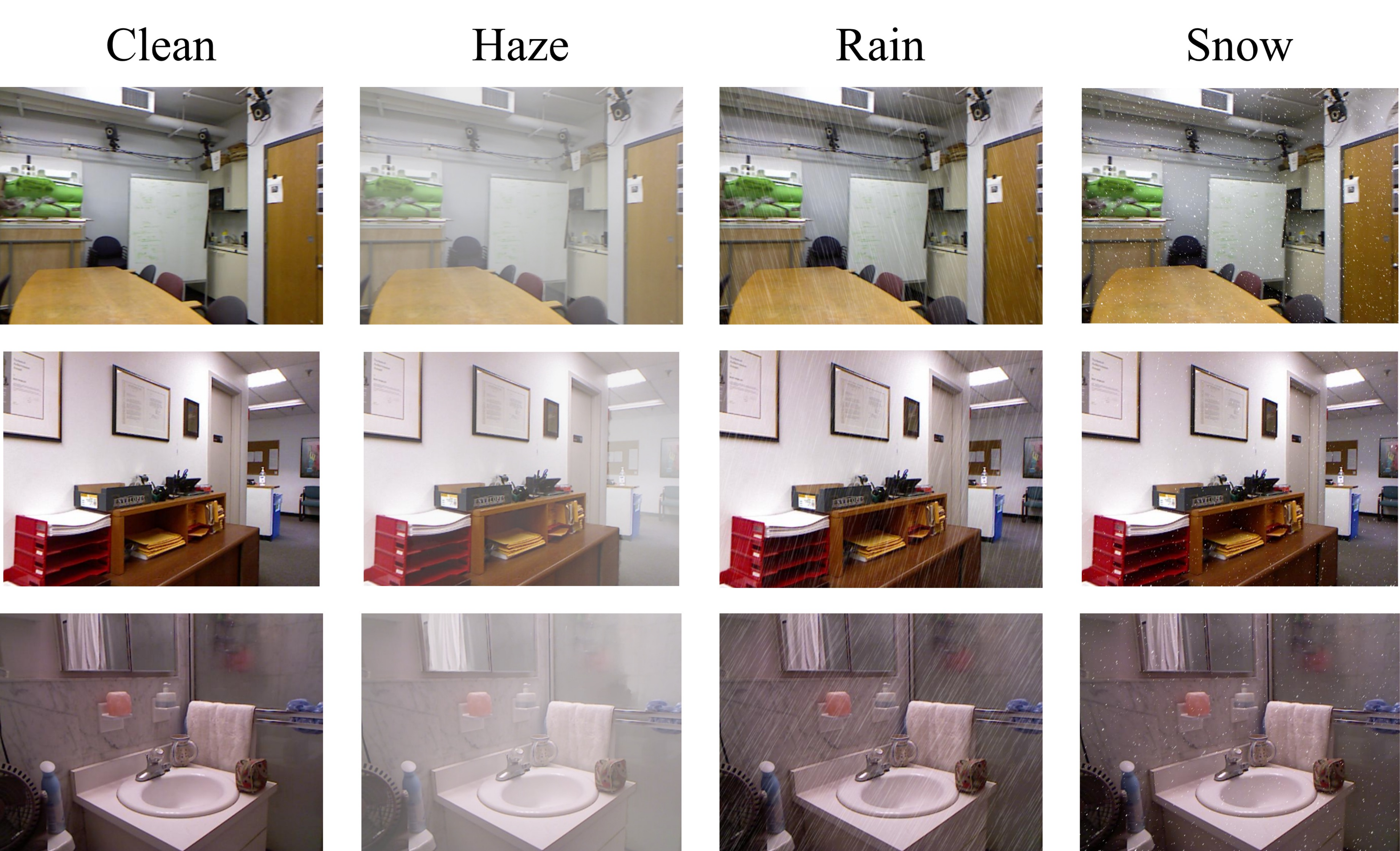}
    \caption{Illustrations of indoor scenes in RESIDE+. For each clean image, we synthesize its hazy, rainy and snowy versions, respectively.}
    \label{fig:Indoor}
\end{figure}
\begin{figure}[t!]
    \centering
    \includegraphics[scale=0.13]{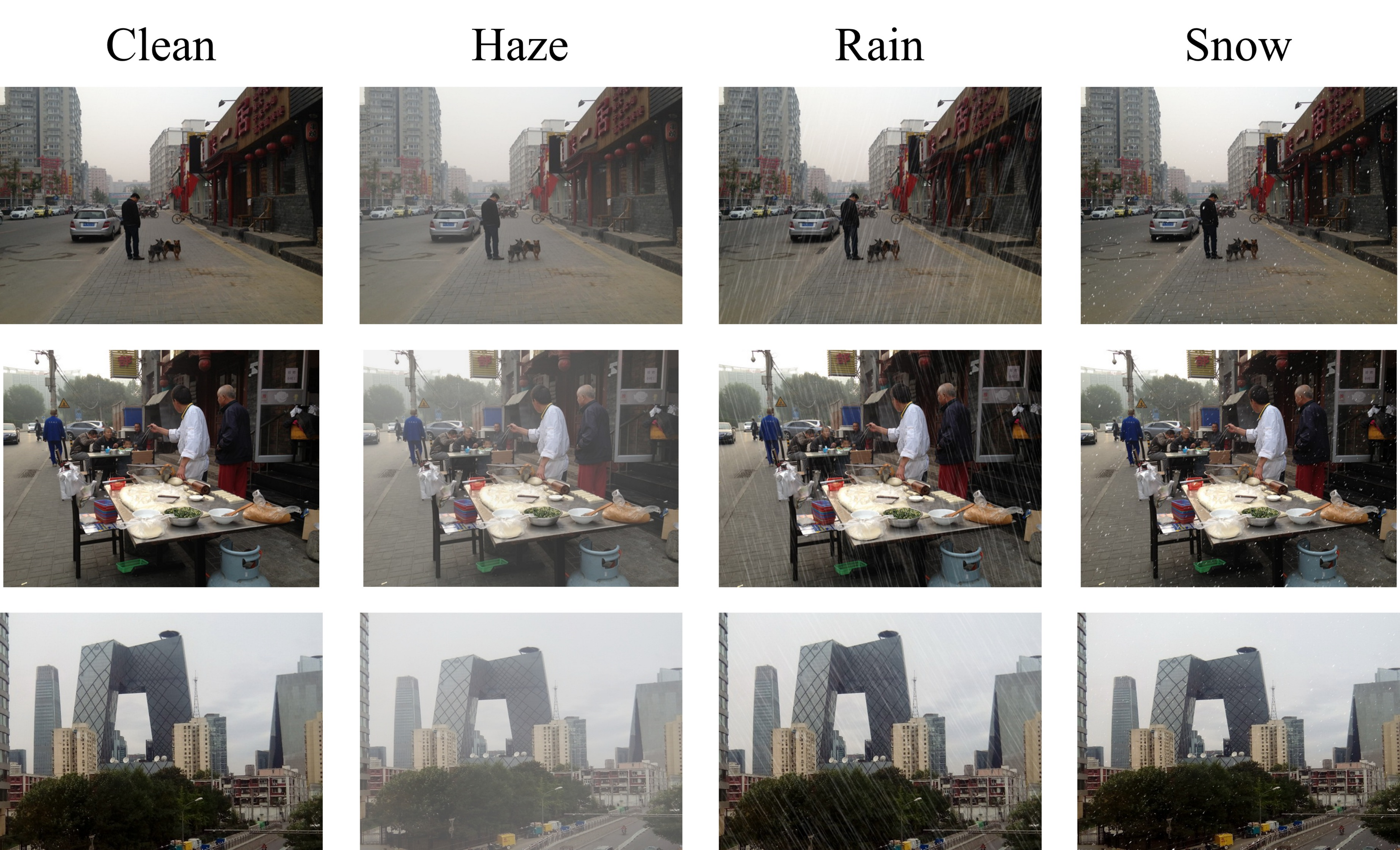}
    \caption{Illustrations of outdoor scenes in RESIDE+. For each clean image, we synthesize its hazy, rainy and snowy versions, respectively.}
    \label{fig:Outdoor}
\end{figure}

Following DDN~\cite{DDN} and DesnowNet~\cite{desnownet}, we adopt PhotoShop to synthesize rainy and snowy images. For rainy images, we sequentially adopt the following steps to obtain the data set, \textit{i.e.}, i) add 150\% uniform noise, ii) conduct Gaussian blur with 0.5 pixel radius, iii) randomly set \textit{angle} as one of $[45, 60, 75, 105, 120, 135]$ and \textit{distance} from 20 to 50 with the gap of 5 to conduct motion blur, and iv) modify the image level. For snowy images, the following steps are taken in sequence, namely, i) add 25\% Gaussian noise, ii) randomly set \textit{angle} as one of $[45, 60, 75, 105, 120, 135]$ and fix \textit{distance} to 3 for conducting motion blur, iii) modify the image level, iv) duplicate and flip the snow layer, v) crystallize the snow with \textit{cell size} from 3 to 9, vi) conduct motion blur again like step ii. Partial indoor and outdoor synthetic examples are shown in \cref{fig:Indoor,fig:Outdoor}, respectively.

\begin{figure}[t!]
    \centering
    \hspace{-1.5em}
    \includegraphics[scale=0.58]{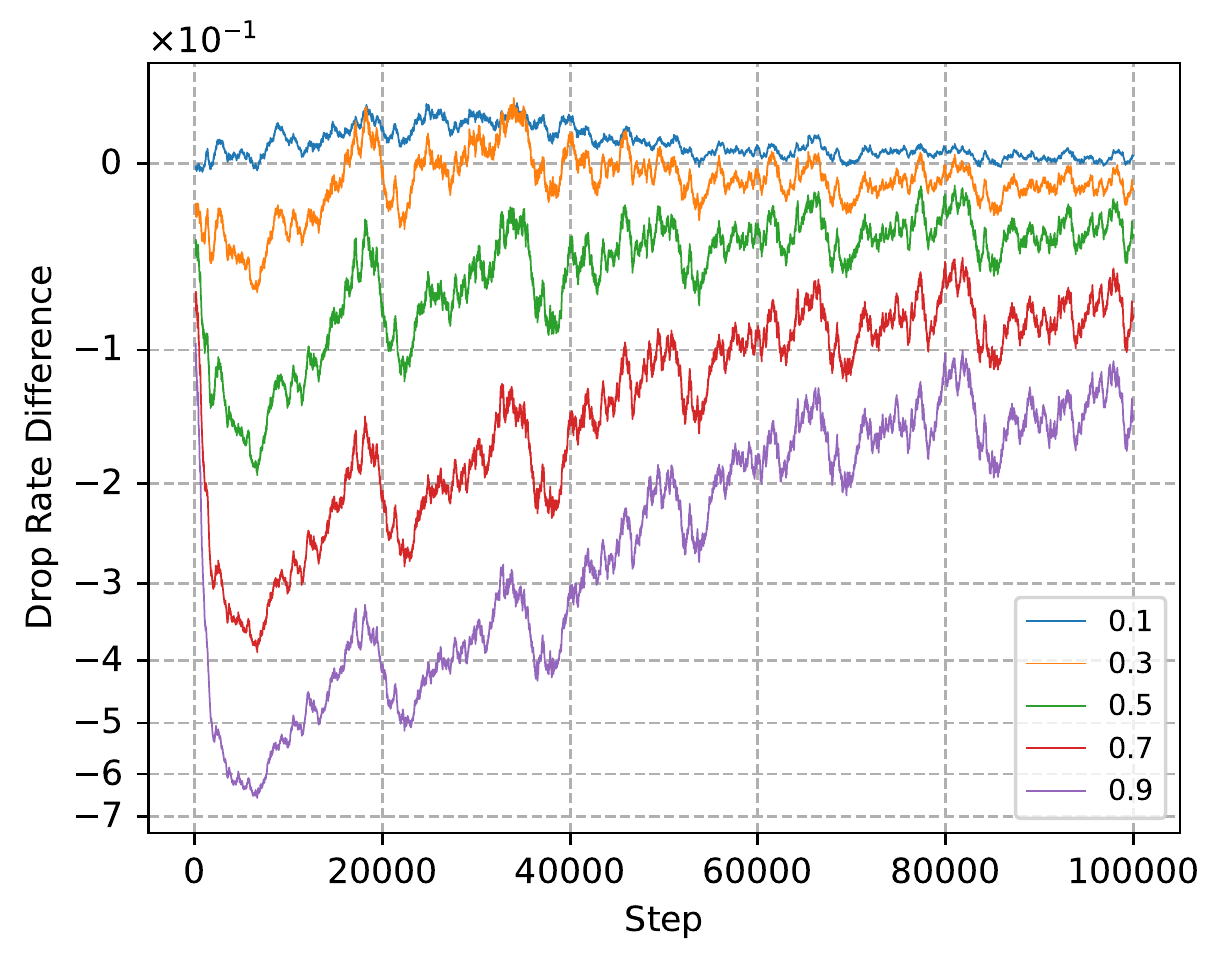}
    \caption{The DRD on different proportions (from 0.1 to 0.9) of auxiliary degradations, \textit{i.e.}, image denoising, during the training stages. With the increment of proportion, the DRD decreases at the same time.}
    \label{Fig:ratio1}
\end{figure}
\section{Experiments}
In this section, we will first introduce the experimental setting and show the qualitative and quantitative results on the proposed RESIDE+. Then, we will make comparisons with task affinity methods and show several intuitive explanations on the effect of auxiliary degradations. Besides, ablation studies are also conducted on sampling steps and validation loss.
 
\subsection{Experimental Settings}
In this section, we introduce the details of the used dataset, baselines, evaluation metrics, and training details.

{\bf Dataset:} 
In our experiments, we adopt the proposed RESIDE+ for training and evaluations, which contains clean images corresponding to its rainy, hazy and snowy version. For noisy images used in the experiments, we generate noisy images by manually adding white Gaussian noise to clean images with $\sigma=15$. 

{\bf Baselines:}  For comprehensive comparisons, we take experiments on seven representative dehazing methods, \textit{i.e.}, AOD-Net~\cite{AOD-Net},  GDN~\cite{GridDehazeNet}, FFANet~\cite{FFA-Net}, GCANet~\cite{GCA-Net}, MSBDN~\cite{MSBDN}, AECR-Net~\cite{AECR-Net} and Dehamer~\cite{dehamer}. To comprehensively demonstrate the effectiveness of our method, four different training settings are examined, \textit{i.e.}, training baselines with hazy-clean pairs only (termed Original), training baselines with 90\% hazy-clean and 10\% noisy-clean image pairs (termed 10\% noise), training baselines with 90\% hazy-clean and 10\% rainy-clean image pairs (termed 10\% rain) and training baselines with 90\% hazy-clean and 10\% snowy-clean image pairs (termed 10\% snow). 

{\bf Evaluation Metrics:} Following~\cite{AOD-Net, AECR-Net, dehamer}, we adopt the Peak Signal-to-Noise Ratio (PSNR)~\cite{PSNR} and the Structural SIMilarity (SSIM)~\cite{SSIM} as the evaluation metrics to measure image quality. Besides, Pearson correlation coefficient (Pearson’s coefficient) is adopted to measure linear correlation between DRI and final performance. Higher value of these metrics indicates better performance of methods.

{\bf Training details:} We conduct experiments in PyTorch on NVIDIA GeForce RTX 3090 GPUs. For fair comparisons, we use the officially released codes to train the networks if they are publicly available. Except for partial training samples are replaced by other degradations, all the experimental settings are the same as the official settings.

\begin{figure}[t!]
    \centering
    \hspace{-1.5em}
    \includegraphics[scale=0.58]{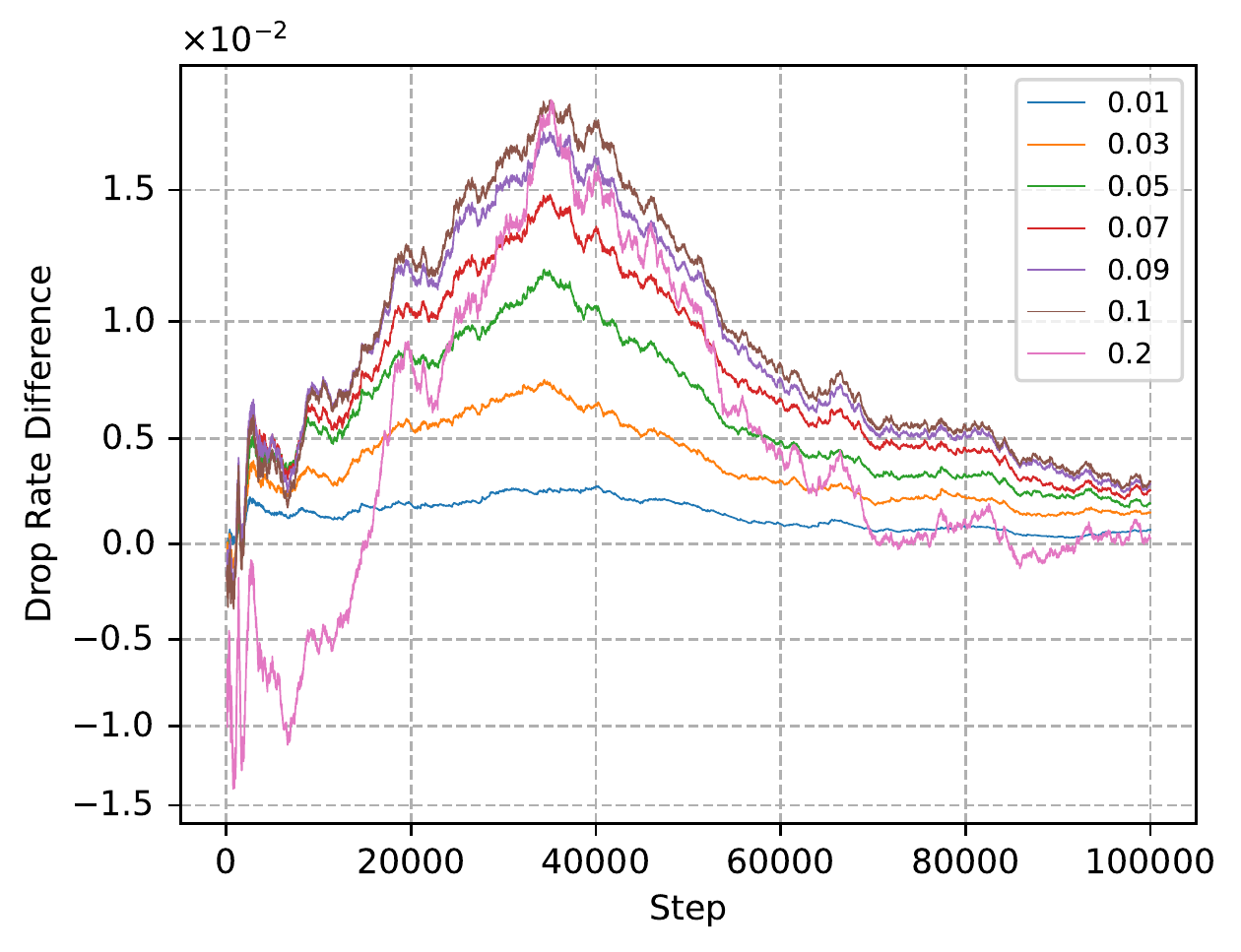}
    \caption{The DRD on different proportions (from 0.01 to 0.2) of auxiliary degradations during the training stages. When the proportion is 0.1, the DRD is the highest.}
    \label{Fig:ratio2}
\end{figure}

\begin{table}[t!]
    \centering
    \small
    \caption{Performances on image dehazing with different proportions of the auxiliary task, \textit{i.e.}, image denoising. Results in boldface indicate the setting with the best performance.}
    \begin{tabular}{c | c c c c c}
        \toprule
        $P$ &    DRI      & PSNR  & $\Delta_{PSNR}$ & SSIM & $\Delta_{SSIM}$  \\
        \midrule
        0\%   &  0        & 33.84 &    -  & 0.9849 & -      \\
        1\%   & 0.00132   & 33.86 &  0.02 & 0.9848 & -0.0001  \\
        3\%   & 0.00353   & 33.87 &  0.03 & 0.9851 & 0.0002 \\
        5\%   & 0.00551   & 33.89 &  0.05 & \bf{0.9859} & \bf{0.0010} \\ 
        7\%   & 0.00704   & 33.92 &  0.08 & 0.9853 & 0.0004 \\
        9\%   & 0.00815   & 33.93 &  0.09 & 0.9853 & 0.0004 \\
        10\%  & \bf{0.00860}  & \bf{33.96} &  \bf{0.12} & 0.9850  &  0.0001\\
        20\%  & 0.00726  & 33.93 & 0.09 & 0.9857  & 0.0008\\
        \bottomrule
    \end{tabular}
    \label{Tab:ratio2}
\end{table}

\subsection{Results on Different Proportions of Auxiliary Degradation} 
\label{Sec:proportion}
In this section, we conduct experiments to study the impact of proportion in degradation combination during the training process. Similar to experiments in~\cref{Sec:DPD}, we take haze as the anchor degradation and noise as the auxiliary degradation to investigate how the proportion influences the performance on the anchor degradation. In detail, the proportions of auxiliary degradation vary from 0.1 to 0.9 with a gap of 0.2.

As presented in~\cref{Fig:ratio1}, we find out that DRI is sensitive to the proportion of auxiliary degradation. In other words, only the degradation combinations with proper proportions of the auxiliary task tend to achieve better performance. Similar conclusions could be derived from the proportions around 10\%, which is shown in~\cref{Fig:ratio2}.

To further explore the influence of the auxiliary degradations' proportions, we also calculate the DRI around the $10\%$ proportion. As shown in~\cref{Tab:ratio2}, we further verify the conclusions mentioned in~\cref{Sec:DPD}, \textit{i.e.}, a positive DRI always indicates the performance improvement on anchor restoration.

\begin{figure*}[ht!]
  \centering
  \hspace{-1.8em}
   \includegraphics[scale=0.138]{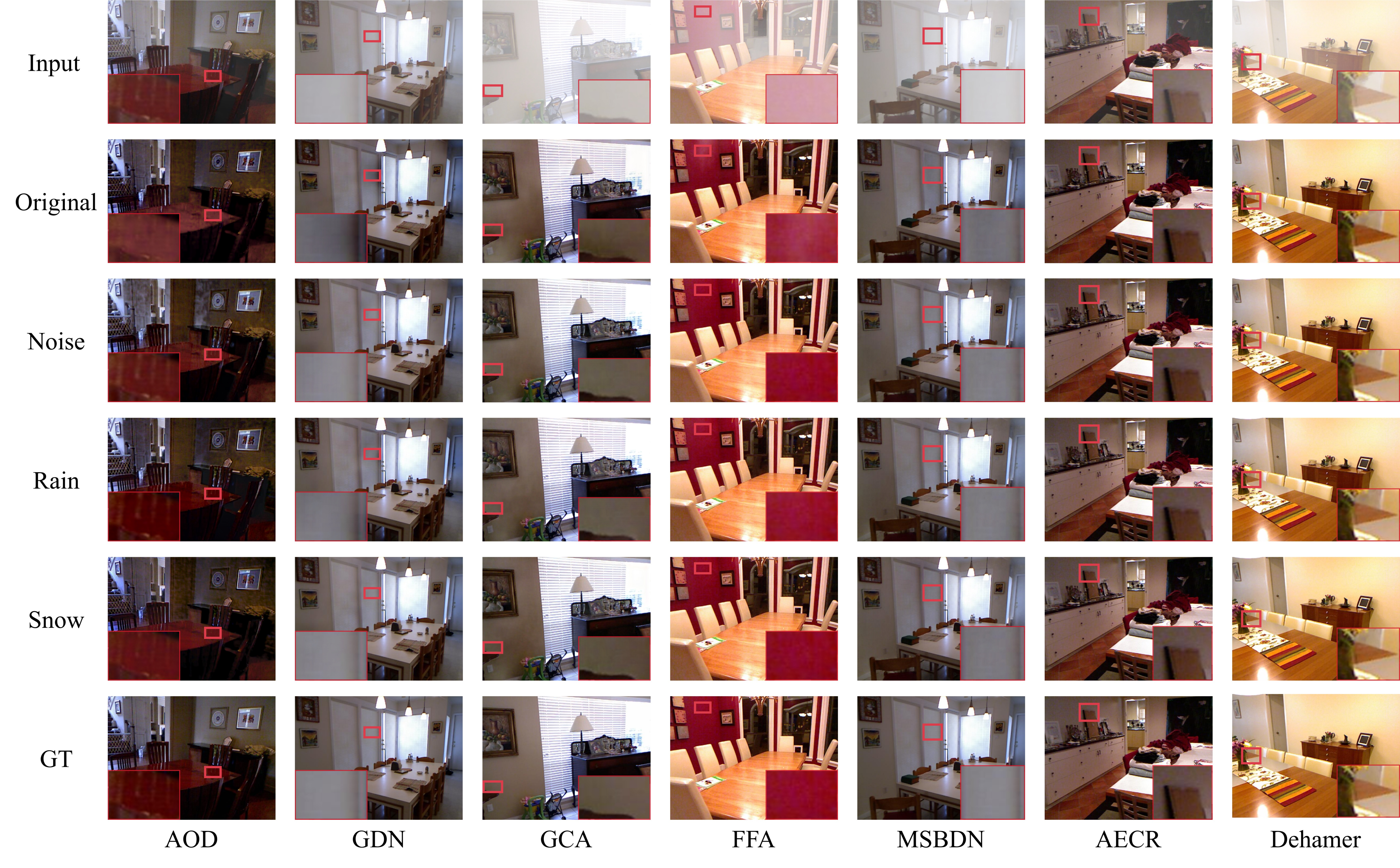}
   \caption{Visual comparisons on indoor scene of SOTS dataset. Some areas are highlighted in red rectangles and zooming-in is recommended for better visualizations and comparisons.}
   \label{Fig:ITS}
\end{figure*}

\begin{table*}[t!]
	\centering
	\footnotesize
	\caption{Quantitative results of image dehazing on SOTS dataset. Performance improvements have been achieved by adopting appropriate degradation Combinations}
	\begin{tabular}{c c | c c c c c c c}
		\toprule
         Scenes
         &
         Settings
		 & AOD-Net~\cite{AOD-Net} 
		 & GDN~\cite{GridDehazeNet}
		 & FFANet~\cite{FFA-Net} 
		 & GCANet~\cite{GCA-Net}
		 & MSBDN~\cite{MSBDN}
		 & AECR-Net~\cite{AECR-Net}
		 & Dehamer~\cite{dehamer} \\
		 \midrule
		 \multirow{4}{*}{Indoor} 
		 & Original       & 19.54/0.8244 & 32.16/0.9836 & 36.39/0.9886 
		 & 30.08/0.9601 & 32.72/0.9806 & 37.17/0.9901 & 36.63/0.9881 \\
		 & 10\% Noise        & 19.73/0.8343 & 32.30/0.9843 & 36.79/0.9866
		 & 30.53/0.9625 & 34.48/0.9837 & 37.28/0.9901 & 36.68/0.9904 \\
		 & 10\% Rain         & 19.99/0.8377 & 32.87/0.9856 & 36.96/0.9893 
		 & 31.24/0.9653 & 35.14/0.9858 & 37.19/0.9902 & 36.87/0.9891 \\
		 & 10\% Snow         & 19.82/0.8310 & 32.67/0.9843 & 36.93/0.9894
		 & 30.72/0.9648 & 34.53/0.9858 & 37.29/0.9903 & 36.79/0.9901 \\
		 \midrule
		 \multirow{4}{*}{Outdoor} 
		 & Original       & 23.52/0.9183 & 30.86/0.9819 & 33.38/0.9840 
		 & 26.08/0.9614 & 33.84/0.9849 & 33.52/0.9840 & 35.18/0.9860 \\
		 & 10\% Noise        & 23.68/0.9193 & 31.10/0.9828 & 33.90/0.9842 
		 & 27.05/0.9648 & 33.96/0.9850 & 33.76/0.9849 & 35.43/0.9860 \\
		 & 10\% Rain         & 24.17/0.9220 & 31.29/0.9835 & 33.97/0.9842 
		 & 27.23/0.9631 & 34.76/0.9857 & 33.81/0.9846 & 36.11/0.9869 \\
		 & 10\% Snow         & 23.99/0.9203 & 31.18/0.9832 & 34.09/0.9847
		 & 27.10/0.9635 & 34.50/0.9851 & 34.14/0.9853 & 35.88/0.9869 \\
		 \bottomrule
	\end{tabular}
	\label{Tab:Quantitative}
\end{table*}

\subsection{Results on Different Auxiliary Degradations}
In this section, we verify the effectiveness of proposed the method on different auxiliary degradations, \textit{i.e.}, noise, rain streak and snow. Thanks to DPD, we find that above degradations could improve the performance on anchor degradation with the proportion of 0.1. Then, we conduct both quantitative and qualitative comparisons on seven representative baselines to demonstrate the effectiveness of the proposed method. Besides, qualitative experiments are conducted.

For quantitative comparisons, as shown in~\cref{Tab:Quantitative}, remarkable improvements have been achieved by elaborately introducing the auxiliary degradations. 
Specifically, baselines training with the help of image denoising, deraining, and desnowing is $0.42$dB, $0.77$dB, and $0.63$dB higher than the original models in terms of PSNR averagely, and $0.0016$, $0.0027$ and $0.0021$ higher in terms of SSIM as well.

\begin{figure*}[t!]
  \centering
   \hspace{-1.8em}
   \includegraphics[scale=0.138]{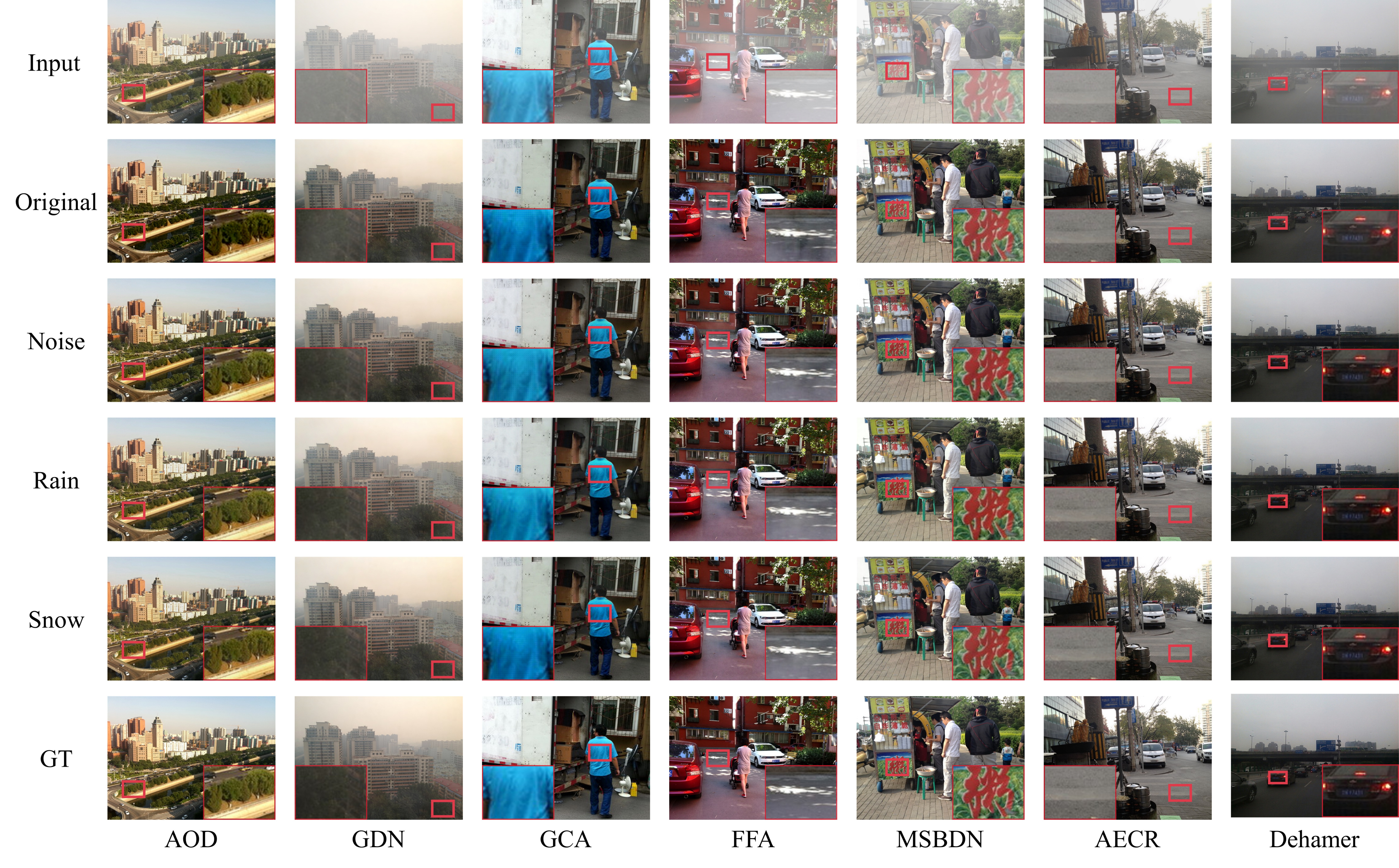}
   \caption{Visual comparisons on outdoor scene of SOTS dataset. Some areas are highlighted in red rectangles and zooming-in is recommended for better visualizations and comparisons.}
   \label{Fig:OTS}
\end{figure*}

For qualitative comparisons, we show visual results on the SOTS testing set with both indoor and outdoor scenes as presented in~\cref{Fig:ITS,Fig:OTS}, respectively. For indoor scenes, as shown in~\cref{Fig:ITS}, we could find out that baselines helped with the auxiliary degradations show better visual results, while original models usually suffer from color distortions. For complex outdoor scenes, the proposed methods still get better visual results. As shown in~\cref{Fig:OTS}, results of our methods are closer to the ground-truth than original ones.

Besides, we conduct extra experiments to investigate how DRD changes during the training process for those degradation combinations. As shown in~\cref{fig:dehaze}, we could see that DRD are positive for those combinations with performance improvement after a few training steps. Quantitative results are shown in the next section.

\subsection{Results on Different Anchor Degradations}
\begin{figure*}
    \begin{center}
    \subfloat[Anchor degradation: {\bf noise}.]{
        \label{fig:denoise}
        \includegraphics[scale=0.6]{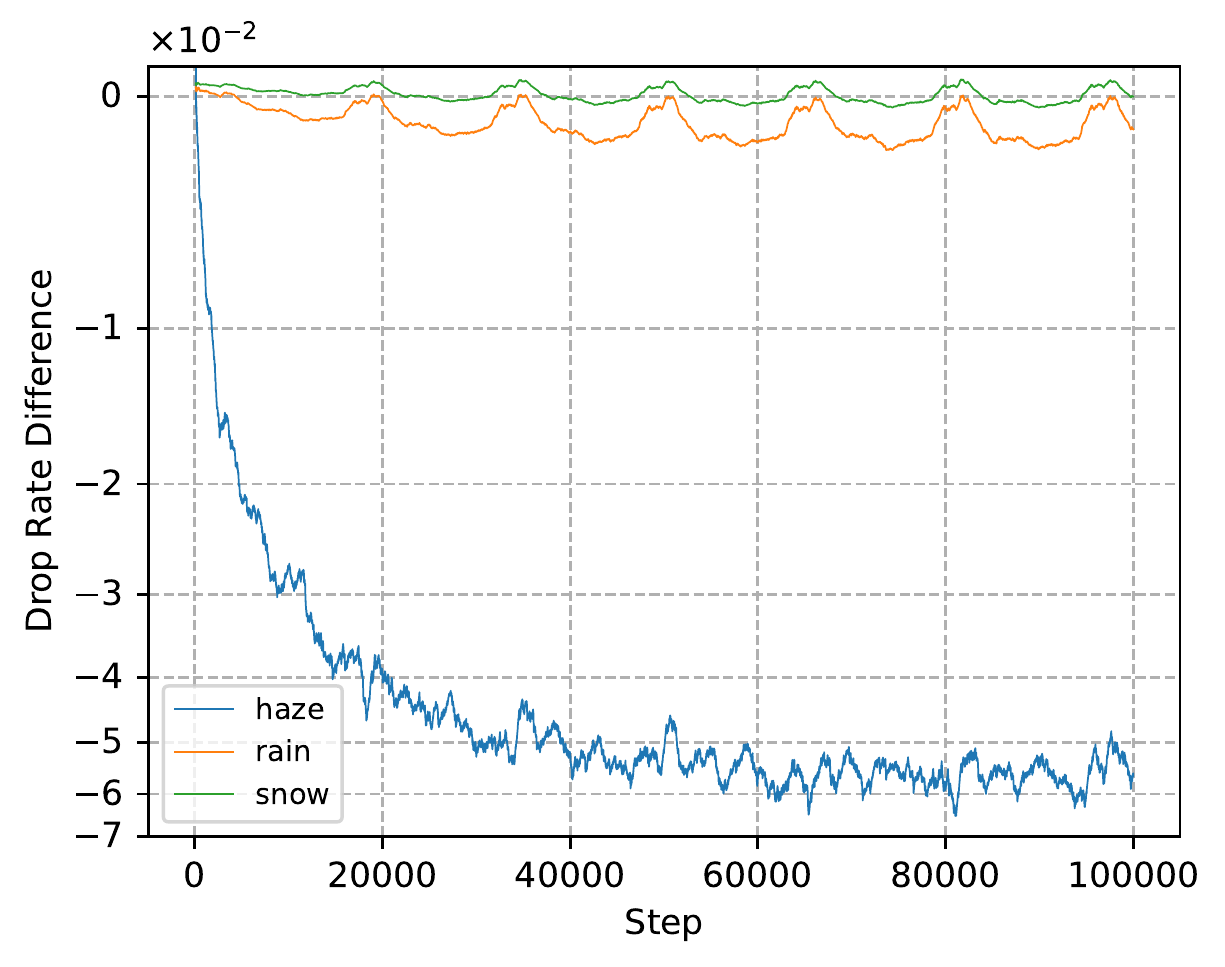}
    }
    \subfloat[Anchor degradation: {\bf rain streak}.]{
        \includegraphics[scale=0.6]{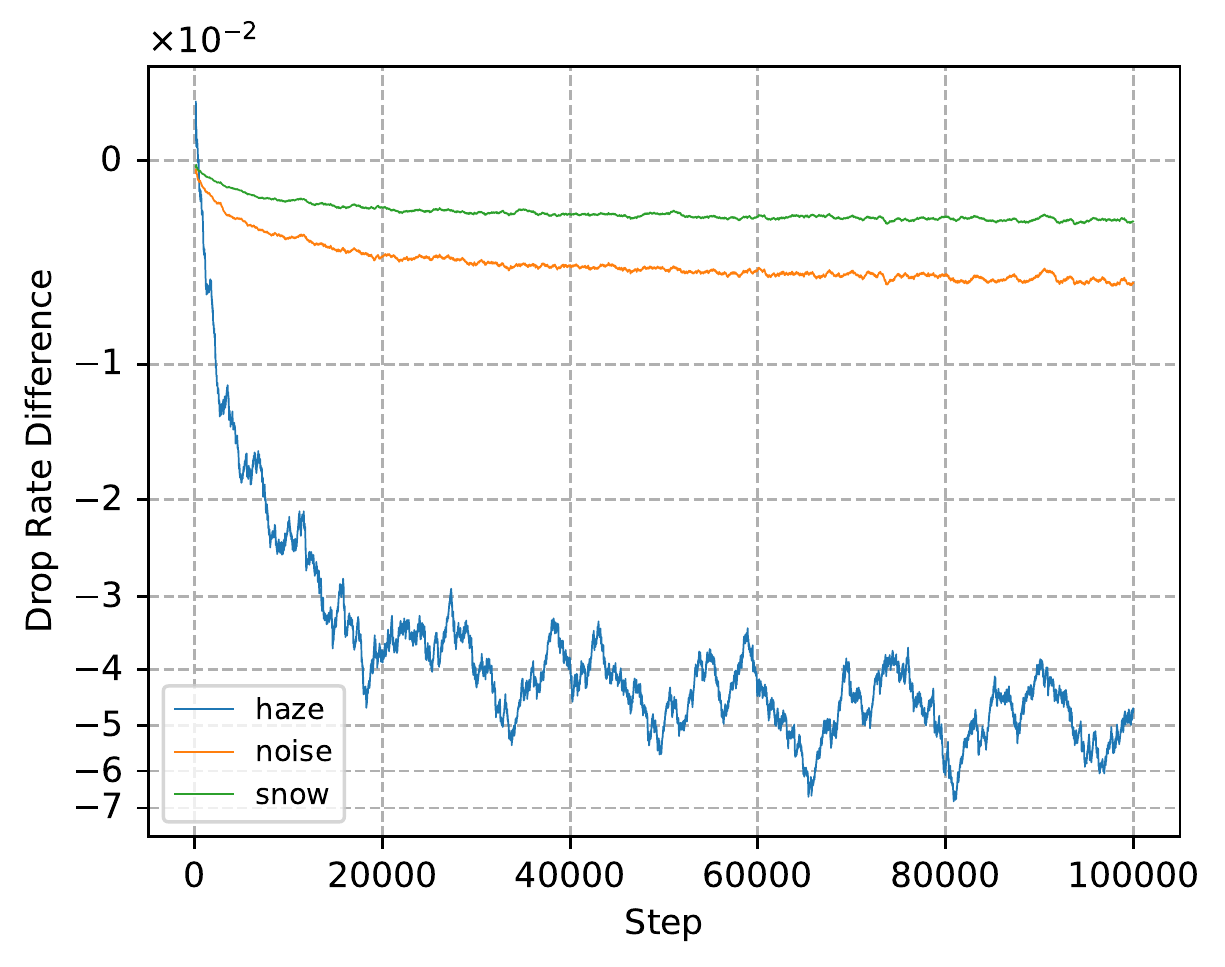}
        \label{fig:derain}
    }

    \subfloat[Anchor degradation: {\bf haze}.]{        \includegraphics[scale=0.6]{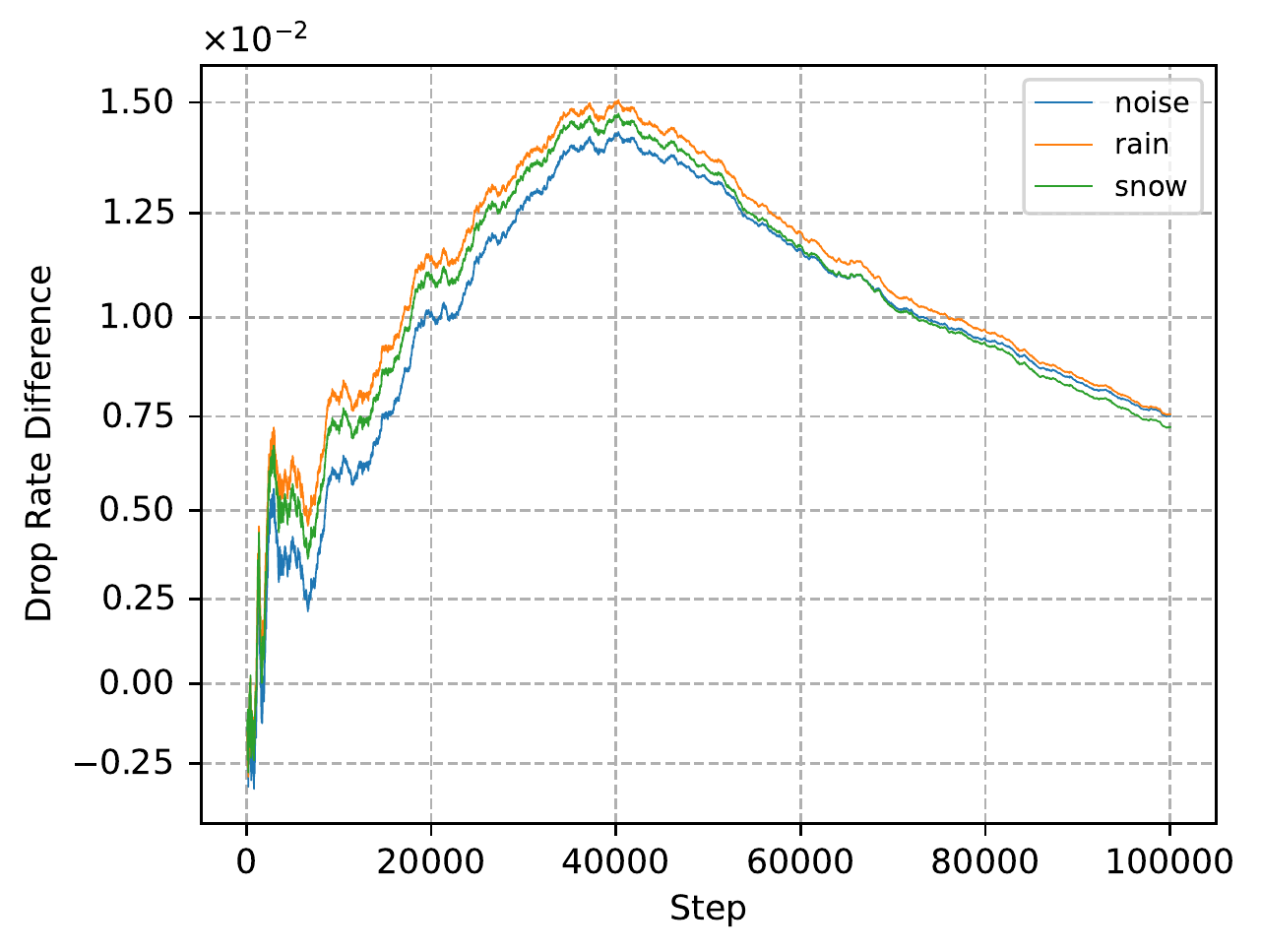}
        \label{fig:dehaze}
    }
    \subfloat[Anchor degradation: {\bf snow}.]{
        \includegraphics[scale=0.6]{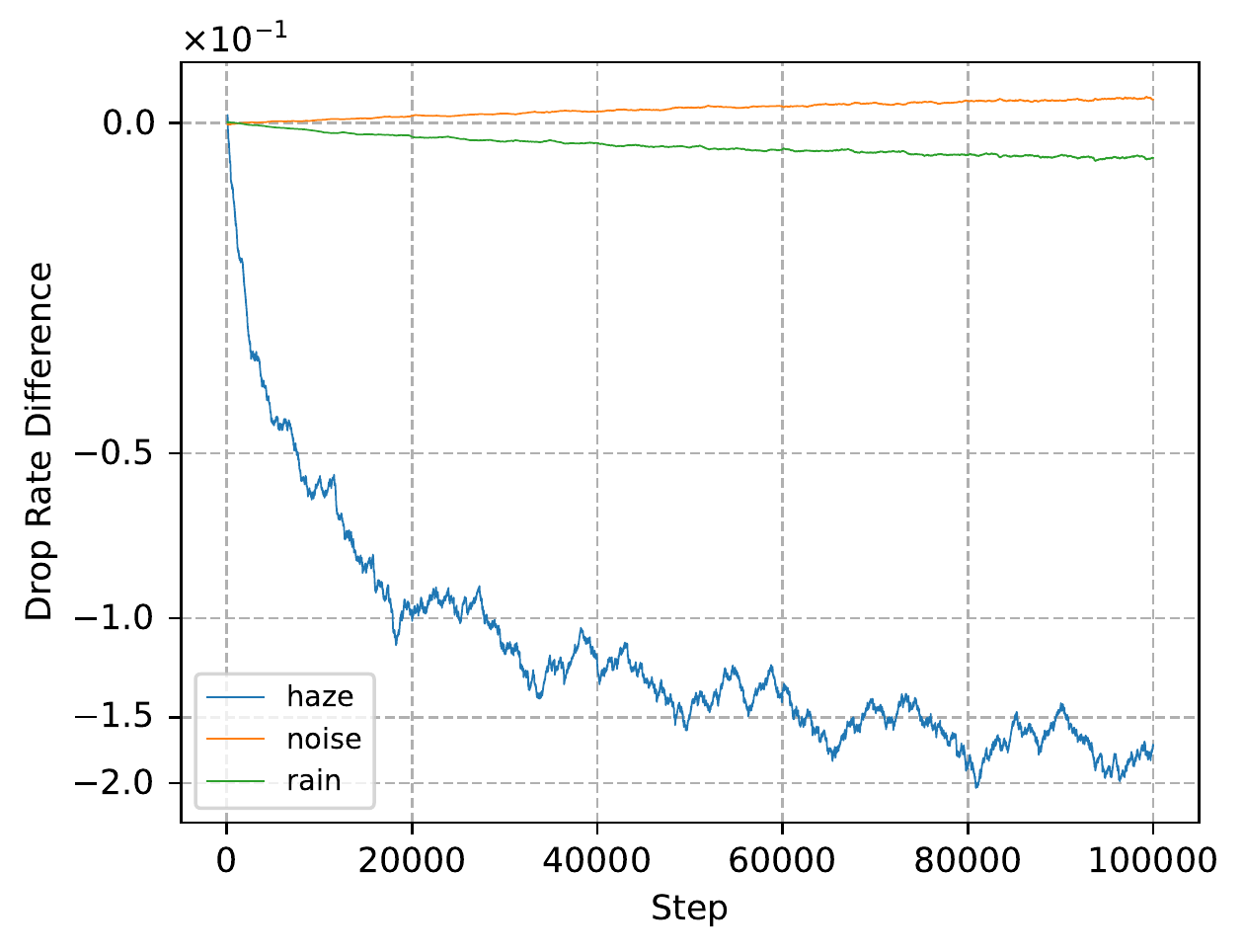}
        \label{fig:desnow}
    }
    \end{center}
    \caption{\label{Fig:Anchor} The DRD on four anchor degradations (\textit{i.e.}, noise, rain streak, haze, and snow) during the training stage. Different colors indicate different auxiliary degradations.}
\end{figure*}

To verify the generalization of the proposed method, we retrain the MSBDN with different combinations of anchor and auxiliary degradations. From results shown in~\cref{Tab:anchor}, we could observe that DRI could be generalized to other anchor degradations and embraces high predictability, \textit{i.e.}, positive DRI indicates the improvement on anchor degradations and vice versa.
As the results present in~\cref{Fig:Anchor}, we find that the DRD is not immutable during the training steps. For instance, as shown in~\cref{fig:denoise}, although the DRI of snow is positive, DRD is sometimes negative.

\begin{table}[t!]
    \centering
    \small
    \caption{Quantitative results of different anchor degradations on the SOTS+ dataset. }
    \begin{tabular}{p{4em}<{\centering}|c| c c c c c }
    \toprule
        Anchor & Metric & Haze & Noise  & Rain  & Snow \\
        \midrule
        \multirow{3}{*}{Original} 
        & PSNR & 33.84  & 34.85  & 38.12   & 36.81 \\
        & SSIM & 0.9849 & 0.9468 & 0.9652  & 0.9778 \\
        & DRI  & 0      & 0      & 0       & 0 \\
        \hline
        \multirow{3}{*}{10\% Haze}
        & PSNR & \multirow{3}{*}{-}  & 34.78     & 36.66     & 36.52 \\
        & SSIM &                     & 0.9457    & 0.9587    & 0.9768 \\
        & DRI  &                     & -0.04845  & -0.04090  & -0.12649 \\
        \hline
        \multirow{3}{*}{10\% Noise}
        & PSNR & 33.96     & \multirow{3}{*}{-} & 37.59    & 36.84  \\
        & SSIM & 0.9850    &                    & 0.9636   & 0.9781 \\
        & DRI  & 0.00860   &                    & -0.00457 & 0.00143 \\
        \hline
        \multirow{3}{*}{10\% Rain}
        & PSNR & 34.76   & 34.83    & \multirow{3}{*}{-}  & 36.80  \\ 
        & SSIM & 0.9857  & 0.9462   &                     & 0.9781 \\
        & DRI  & 0.00879 & -0.00097 &                     & -0.00234 \\
        \hline
        \multirow{3}{*}{10\% Snow} 
        & PSNR & 34.50   & 34.89   & 37.61    & \multirow{3}{*}{-}  \\
        & SSIM & 0.9851  & 0.9470  & 0.9635   &                     \\
        & DRI  & 0.00865 & 0.00024 & -0.00167 & \\
        \bottomrule
    \end{tabular}
    \label{Tab:anchor}
\end{table}

\subsection{Comparison with Task Affinity Methods}
\label{Sec:TA}
In this section, we take haze as the anchor degradation, $10\%$ noise, $10\%$ rain, and $10\%$ snow as the auxiliary degradation to make comparisons with two representative task affinity methods, \textit{i.e.}, Taskonomy~\cite{Taskonomy} and TAG~\cite{TAG}. 
As shown in \cref{Tab:affinity}, our DRI embraces higher predictability. In brief, taking $10\%$ noise as an example, the Taskonomy and TAG affinity are $-1.00000$ and $-0.00189$, which goes against the final performance on anchor degradation. While, the DRI is $0.00860$, corresponding with the performance improvement. 

\begin{table}[t]
    \centering
    \footnotesize
    \caption{The degradation affinity results of different metrics.}
    \begin{tabular}{c| c c c c}
        \toprule
        Settings   & Original & $10\%$ Noise  & $10\%$ Rain   & $10\%$ Snow \\
        \midrule
         PSNR & 33.84  & 33.96  & 34.76  & 34.50 \\
         SSIM & 0.9849 & 0.9850 & 0.9857 & 0.9851 \\
        \hline
        Taskonomy~\cite{Taskonomy}  & -
                              & -1.00000 & -0.00215   & -0.000005\\
        TAG~\cite{TAG}              & -     
                              & -0.00189 & -0.00060   & -0.00142\\
        DRI                         & -
                              & 0.00860 & 0.00879 & 0.00865\\
        \bottomrule
    \end{tabular}
    \label{Tab:affinity}
\end{table}

\begin{figure}[t]
    \centering
    \includegraphics[scale=0.48]{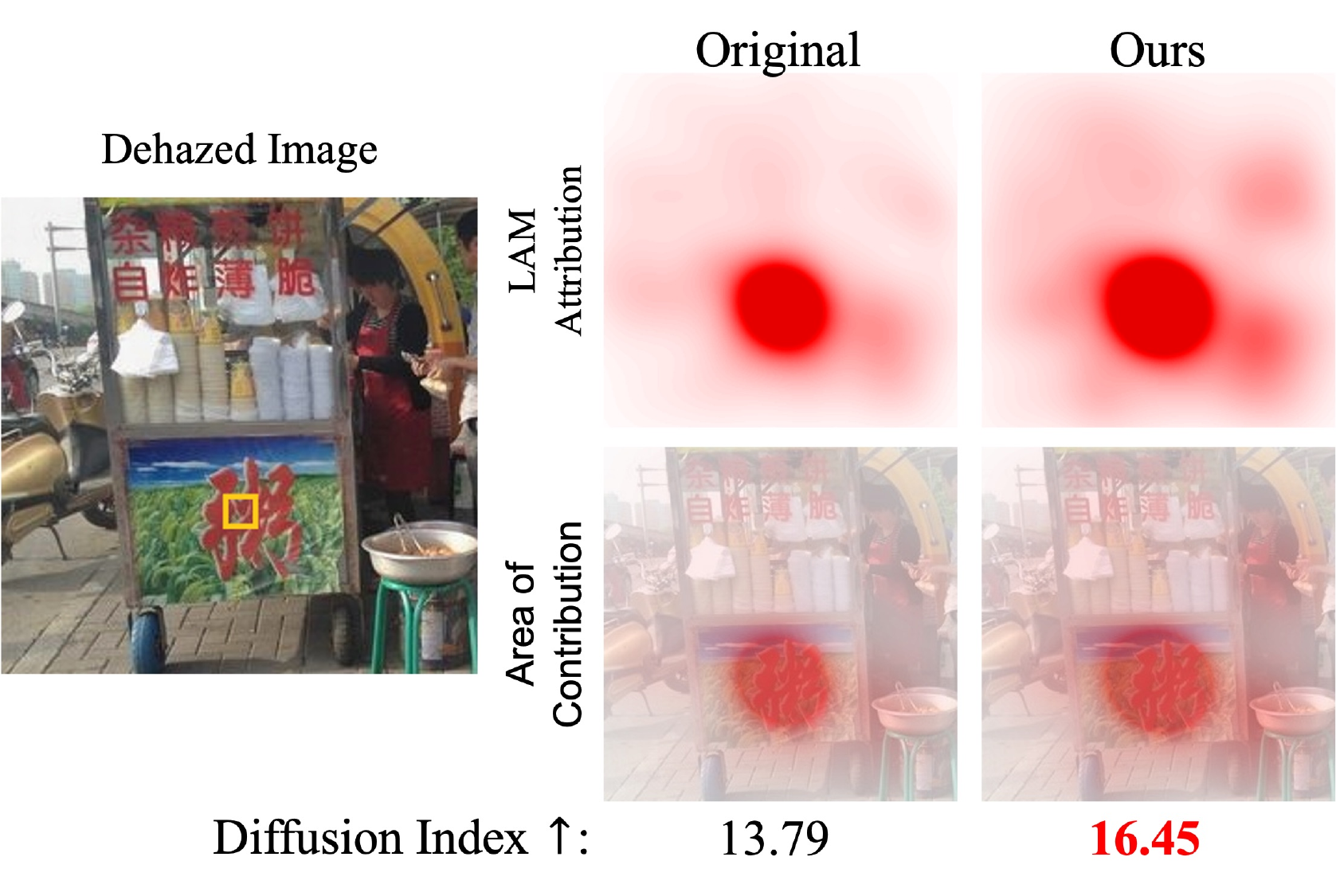}
	\caption{LAM and DI for dehazing model interpretation.}
	\label{Fig:LAM}
\end{figure}

\subsection{Intuitive Explanation on Auxiliary Degradations}
To further investigate and provide an intuitive explanation on how the auxiliary degradations boost the performance on anchor degradations, we adopt the Local Attribution Maps (LAM)~\cite{LAM} to analyze how the restoration network leverages pixel information from inputs for recovering given local regions. As shown in~\cref{Fig:LAM}, one could find out that the auxiliary degradation will enable the network to utilize more information for recovery (see yellow rectangle). Furthermore, Diffusion Index (DI)~\cite{LAM} is a metric that quantifies how much information the network leverages from the input image. A larger DI indicates more pixels are involved, and our method surpasses the original model, which verifies that auxiliary degradation enlarges the influence areas of the original network.

\subsection{Ablation Studies}
In this section, we conduct further explorations on DRI. We first compute the DRI with different sampling steps to searching an optimal step that could balance between high performance and low computational cost. Then, instead of calculating the validation loss, we calculate DRI with training loss to investigate the necessity of validation loss.

\begin{figure}[t]
    \centering
    \hspace{-1em}
    \includegraphics[scale=0.58]{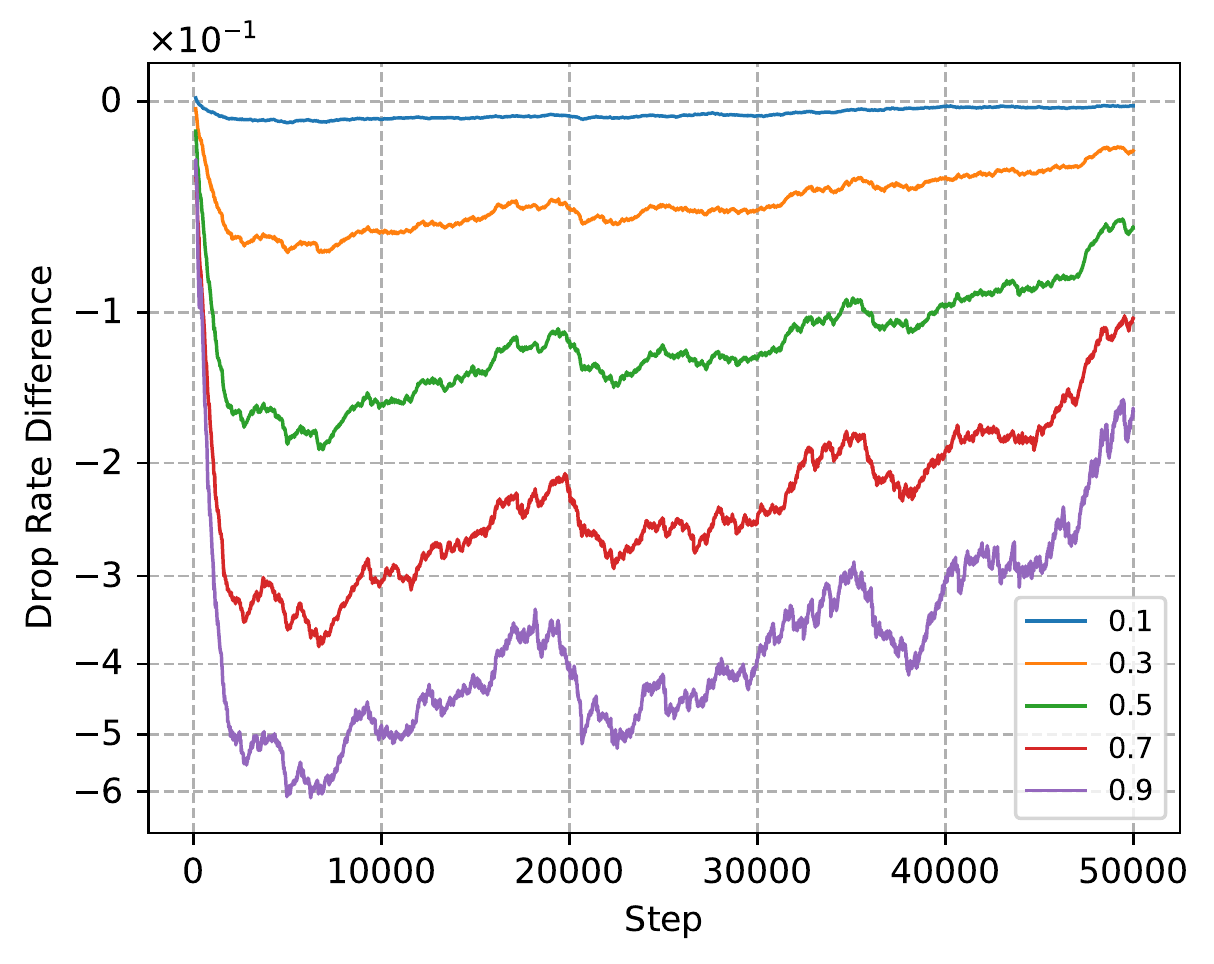}
    \caption{DRD calculated on training loss during the training stage. Different colors indicate different proportions of auxiliary degradations.}
    \label{Fig:TrainDRIRatio}
\end{figure}

\begin{table}[t!]
	\centering
	\footnotesize
	\caption{Quantitative results on different sampling steps. Speedup is relative to computing DRI in every 1 step. The best results on the Pearson coefficient and the speedup are shown in boldface.}
	\begin{tabular}{c | c | c }
		\toprule
		 Method  
		 & Pearson's Coefficient
		 & Relative Speedup \\
		 \midrule
		 Every 1 Step  & 0.9698 & 1.0$\times$ \\
		 Every 5 Steps  & \bf{0.9721} & 1.93$\times$ \\
		 Every 10 Steps  & 0.9689 & 2.26$\times$ \\
		 Every 50 Steps  & 0.9586 & 2.45$\times$ \\
		 Every 100 Steps  & 0.9693 & 2.48$\times$ \\
		 First 30\% Steps  & 0.9718 & \bf{3.33$\times$} \\
		 Middle 30\% Steps  & 0.9706 & 2.29$\times$ \\
		 Final 30\% Steps  & 0.9674 & 1.73$\times$ \\
		 \bottomrule
	\end{tabular}
	\label{Ablation:Step}
\end{table}

\subsubsection{Ablation Study on Different Steps}
\label{Sec:step}
Calculating DRI with every step's DRD is somehow time-consuming, it is highly expected to find a way to make a balance between performance and computational cost. Inspired by the former experimental results, we find out that the DRD value in consecutive steps is likely to be similar. It indicates that we could calculate DRI with a large sampling step instead of every step, which would be time-saving. To this end, we calculate the DRI for every $1$, $5$, $10$, $50$, $100$ steps, and evaluate the correlation between DRI and final performance on anchor degradations by Pearson’s coefficient. Besides, we also evaluate DRI in the first, middle, and final 30\% steps to verify its performance. As shown in~\cref{Ablation:Step}, we could observe that a large number of sampling intervals are not necessary to predict the final performance. In other words, we could only compute DRI within a few steps, \textit{e.g.}, first $30\%$ steps, and obtain convincing results. 

\begin{figure}[t]
    \centering
    \hspace{-1em}
    \includegraphics[scale=0.58]{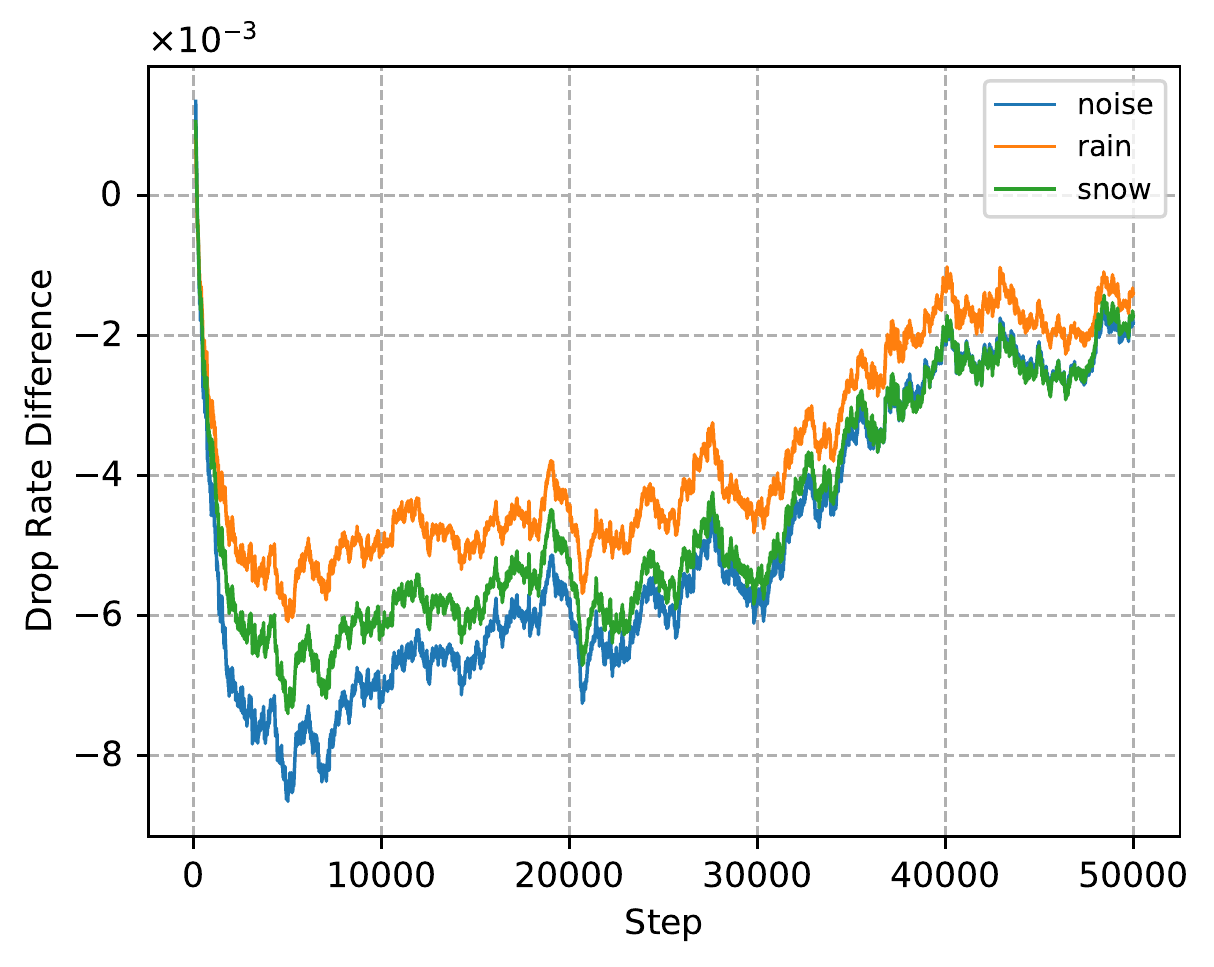}
    \caption{DRD is calculated on training loss with different degradation combinations during the training stage. Different colors indicate different auxiliary degradations.}
    \label{Fig:TrainDRIType}
\end{figure}

\subsubsection{Ablation Study on Validation Loss}
\label{Sec:Val}
In this section, we investigate the necessity of validation loss by calculating DRI on the training set instead. Different from validation loss, the training loss cannot reflect the impact of auxiliary degradation as shown in~\cref{Fig:TrainDRIRatio,Fig:TrainDRIType}. In detail, as shown in~\cref{Fig:TrainDRIRatio}, all DRI are negative during the training process, which go against the observations and results shown in~\cref{Fig:observation,Fig:ratio1,Fig:ratio2}. Taking proportion of 0.1 as an example, it leads to performance improvement as shown in~\cref{Tab:ratio1,Tab:ratio2}, while its DRI calculating on training loss is negative, which indicates the unavailability of training loss. Similar conclusions could be derived from~\cref{Fig:TrainDRIType}, we can easily observe that the DRI of different auxiliary degradations are negative according to DRD in the training process, which goes against their performance. Hence, two conclusions could be derived, i) training an anchor restoration only could converge fast on the training set, while the model trained with both anchor and auxiliary could achieve lower validation loss. In other words, appropriate auxiliary degradation could improve the generalization ability of the model; ii) Instead of resorting to achieving lower training loss value, less-touched validation loss should be paid more attention in the future, which is highly related to the final performance. 

\section{Conclusions}
In this paper, we propose the Degradation Relationship Index (DRI) to quantify the degradation relationship through measuring the mean drop rate difference of validation loss between two models, where one is trained with the anchor degradation and the other is trained with plus the auxiliary degradation. Thanks to DRI, one could improve the restoration performance of the anchor task by introducing a specific auxiliary degradation with the desirable degradation proportion. Besides, a novel dataset (termed RESIDE+) is constructed to eliminate the content discrepancy as a benchmark for this new topic. Extensive experiments verify the effectiveness of them.

It should be pointed out that this work is just an initial attempt to exploration and exploitation on the relationship quantification of image degradations, which might be improved from the following aspects. On the one hand, it is worthy to further explore how to directly quantify the ``similarity'' instead of only the influence between degradations. Furthermore, it is also promising to understand the relationship between the ``similarity'' and the influence studied in this paper. On the other hand, our DPD takes an intuitive but effective way to determine the proportion of the auxiliary degradation, which might be improved through automatic machine learning methods in the future. 

{\small
\bibliographystyle{ieee_fullname}
\bibliography{egbib}
}

\end{document}